\documentclass[10pt,journal, final]{IEEEtran}
\usepackage{ifpdf}
\usepackage{cite}
\usepackage{algorithmic}
\usepackage{algorithm}
\ifCLASSINFOpdf
\usepackage[pdftex]{graphicx}

\else
 \usepackage[dvips]{graphicx}
\fi
\usepackage[cmex10]{amsmath}
\usepackage{epstopdf}
\usepackage{array}
\usepackage{flushend}
\usepackage{balance}
\usepackage{eqparbox}

\usepackage[tight,footnotesize]{subfigure}
\usepackage{fixltx2e}
\usepackage{color}
\usepackage{float}

\usepackage{dblfloatfix}
\usepackage{url}
\usepackage{cite}
\usepackage{subfigure}
\usepackage{amsmath}
\usepackage{epstopdf}

\usepackage{amssymb}
\begin{document}
\title{Machine Learning-Based User Scheduling in Integrated Satellite-HAPS-Ground Networks}
\setlength{\columnsep}{0.21 in}
\author{Hayssam~Dahrouj,~\IEEEmembership{Senior Member,~IEEE,}
        Shasha~Liu,~\IEEEmembership{Student Member,~IEEE,}
        and
        Mohamed-Slim~Alouini,~\IEEEmembership{Fellow,~IEEE}

\thanks {This paper is accepted for publication at the IEEE Network Magazine.
Hayssam Dahrouj is with the department of Electrical Engineering, University of Sharjah, Sharjah, UAE (e-mail: hayssam.dahrouj@gmail.com).

Shasha Liu and Mohamed-Slim Alouini are with the Division of Computer, Electrical and Mathematical Sciences and Engineering, King Abdullah University of Science and Technology, Thuwal 23955-6900, Saudi Arabia (e-mail: shasha.liu@kaust.edu.sa, slim.alouini@kaust.edu.sa).

}
}
\maketitle
\vspace{-0.8cm}
\begin{abstract}
Integrated space-air-ground networks promise to offer a valuable solution space for empowering the sixth generation of communication networks (6G), particularly in the context of connecting the unconnected and ultraconnecting the connected.
Such digital inclusion thrive makes resource management problems, especially those accounting for load-balancing considerations, of particular interest. The conventional model-based optimization methods, however, often fail to meet the real-time processing and quality-of-service needs, due to the high heterogeneity of the space-air-ground networks, and the typical complexity of the classical algorithms. {Given the premises of artificial intelligence at automating wireless networks design and the large-scale heterogeneity of non-terrestrial networks, this paper focuses on showcasing the prospects of machine learning in the context of user scheduling in integrated space-air-ground communications}. The paper first overviews the most relevant state-of-the art in the context of machine learning applications to the resource allocation problems, with a dedicated attention to space-air-ground networks. {The paper then proposes, and shows the benefit of, one specific use case that uses ensembling deep neural networks for optimizing the user scheduling policies in integrated space-high altitude platform station (HAPS)-ground networks}. Finally, the paper sheds light on the challenges and open issues that promise to spur the integration of machine learning in space-air-ground networks, namely, online HAPS power adaptation, learning-based channel sensing, data-driven multi-HAPSs resource management, and intelligent flying taxis-empowered systems.
\end{abstract}
\begin{IEEEkeywords}
Machine learning, satellite-HAPS-ground network, resource management, user scheduling, ensembling deep neural networks.
\end{IEEEkeywords}
\IEEEpeerreviewmaketitle
\vspace{-0.8cm}
\section{Introduction}
\subsection{Motivation}
Due to their inherent limitations, the existing 5G cellular networks cannot achieve the full power of its initial premises (i.e., Internet of Everything applications (IoE)), which motivates researchers to explore innovative techniques to further empower the sixth generation of communication networks (6G) \cite{saad2019vision}. In fact, existing communications infrastructures mandate major restructuring to meet the deluge in data demand, so as to face the rapid growth of mobile devices and wireless traffic \cite{saeed2021P2P},\cite{alam2021high}.
6G systems, therefore, aspire to achieve high reliability, low latency, and high data rate to provide mobile broadband reliable low latency communication (MBRLLC), massive ultra-reliable low latency communications (mURLLC), and human-centric services (HCS) \cite{saad2019vision}.

While terrestrial densified networks have their own merits at increasing system throughput \cite{alam2021high}, areas such as isolated mountains, oceans, glaciers, and disasters surfaces, may not be suitable for the excessive deployment of base-stations. Further, ultra-dense networks often prove to be interference-exacerbated, all of which motivate the need to augment the ground communication systems with connectivity from the sky. Current satellite-ground networks provide a degree of seamless connectivity to extreme areas; however, the incurring transmission delay becomes substantial in large-scale network deployment. Embedding additional layers in between the satellite and ground platforms, also known as integrated space-air-ground networks, emerges therefore as a strong enabler to jointly improve the data throughput and achieve the global digital inclusion needed in 6G systems \cite{liu2018space,ShashaTWC2022}. {This paper, therefore, focuses on one particular space-air-ground system model, and showcases how machine learning plays a vital role in optimizing the user scheduling policy of the considered network through ensembling deep neural networks}.

{The projected performance of integrated terrestrial and non-terrestrial networks (TNTNs), often referred to as vertical heterogeneous networks (VHetNets), strongly depends on the nature of the interconnecting segments (i.e., satellite segment, air segment, and ground segment). The joint resource allocation problem across such heterogenous systems becomes, therefore, of considerable complexity, e.g., see \cite{liu2018space, kato2019optimizing,ShashaTWC2022} and references therein, especially when accounting for practical system design constraints and the different modes of intra-layer and inter-layer interference.
For instance, traditional model-based optimization algorithms usually require an accurate communication model, which is impossible to obtain in reality. In addition, thanks to the intricacy of most of the recent works on resource allocation in VHetNets \cite{liu2018space, kato2019optimizing,ShashaTWC2022}, existing solutions remain of heuristic nature, and often exhibit high computational complexity that is incompatible with real-time processing of practical networks. Given the recent advances of machine learning, especially with regard to nonlinear mapping and powerful data mining capabilities, machine learning nowadays emerges as a strong tool to solve challenging resource management and optimization problems in wireless communications \cite{dahrouj2021overview}, especially in the context of VHetNets intricate problems. We next present a concise overview of such works, and motivate for one specific new application that uses ensembling deep neural networks for optimizing the user scheduling policies in integrated space-high altitude platform station (HAPS)-ground networks.}

\subsection{Related Work}
{The machine learning framework illustrated in this paper is related to the general resource allocation problem of wireless networks, its applications to space-air-ground communications, and the recently emerging learning-based optimization techniques in wireless systems. We next classify the state-of-the art of such topics in a systematic fashion, and motivate for the use-case adopted in this paper.}

{The general framework of resource allocation includes optimizing resources (e.g., channels and bandwidth) and computing resources (e.g., power and memory) so as to optimize specific network utiltities. Other dimensionality-dependent resources also include beamforming in multiple-input multiple-output (MIMO) systems,  time-slot allocations in time-division multiple-access system (TDMA), frequency-bin allocations in frequency-division multiple-access system (FDMA), etc. The conventional approach to solve such resource allocation problems is to tackle well-modelled optimization problems using mathematical programming techniques, e.g., \cite{liu2018space, kato2019optimizing,ShashaTWC2022}. Such optimization problems, typically concerned with throughput maximization, transmit power minimization, energy-efficiency maximization, etc., often aim at optimizing the network resources, e.g., user scheduling, power, spectrum optimization, etc. For example, in the particular context of resource allocation in space-air-ground communications, existing works that solve the user scheduling problem mostly focus on throughput maximization through numerical heuristic algorithms, e.g., the recursive shrink-and-realign process in \cite{alsharoa2020improvement}, and the joint integer linear programming, generalized assignment problem, and weighted-minimum mean squared error (WMMSE) in \cite{ShashaTWC2022}}. {Despite the numerical prospects of the above classical techniques, recent references, e.g., \cite{zhu2020application, kato2019optimizing,  sun2018learning,liang2019towards,cui2019spatial,cui2021scalable}, show that machine learning techniques can indeed outperform conventional algorithms. This is especially the case for non-terrestrial networks, where classical optimization become obsolete.}

{The rationale behind the non-practicality of the use of conventional algorithms in VHetNets is twofold. Firstly, VHetNets are often quite complex from an optimization algorithmic perspective \cite{ShashaTWC2022}, which particularly stems from the multiples modes of coupled cross-layer interference (i.e., intra-layer interference, intra-layer interference, inter-base-station interference, and intra-base-station interference). Secondly, the stochasticity of VHetNets underlying channels renders the problem modelling a daunting task, especially given the ever-changing nature of the space-air-ground channels. Strictly speaking, utilizing classical optimization algorithms would imply solving the problems in an offline fashion, and repeating the solution process whenever the cascaded multi-layered channel changes. Such a process is clearly unfeasible for any reasonably sized VHetNet. For instance, paper \cite{zhu2020application} points out that machine learning can be used to solve four main problems in integrated air-space-ground networks (i.e., resource management, security authentication, attack detection, and target recognition and location). Reference \cite{kato2019optimizing} further studies the difficulties of resource optimization due to the heterogeneity of space-air-ground networks and proposes adopting machine leaning to improve traffic control performance in one particular system model. References \cite{zhu2020application,kato2019optimizing}, however, do not consider the user scheduling issues, which the current paper tackles via ensembling deep neural network in a space-HAPS-ground system.}

From methodologies perspectives, references \cite{sun2018learning, liang2019towards} adopt deep learning to optimize power control in general wireless networks setups. Particularly, \cite{sun2018learning} illustrates how a class of signal processing algorithms can be approximated by deep neural network. Reference \cite{liang2019towards}, on the other hand, proposes using ensembling learning to improve the system performance as compared to a single deep neural network.
Similarly, device-to-device (D2D) scheduling problem is solved in \cite{cui2019spatial} using spatial deep learning, whereby spatial convolutional filters are adopted to estimate the interference and channel strength of each link from geographic location information as feature vectors of the subsequent fully connected network.
Further, reference \cite{cui2021scalable} employs scalable reinforcement learning (i.e., deep Q-learning) to jointly optimize the routing and spectrum allocation, where a multi-agent distributed method is adopted to optimize its own flow and deep learning is used to predict the Q-value.

\subsection{Contributions}

{Unlike the aforementioned references, the current paper focuses on one particular VHetNet system model, where one geo-satellite connects to one HAPS platform so as to improve the throughput of the ground-level communications. The performance of the considered network, hereafter denoted by space-HAPS-ground network, becomes a strong function of the ground users' scheduling policy, i.e., user-to-HAPS or user-to-ground base-station association protocols.  The paper then proposes, and shows the benefit of, using ensembling deep neural network for optimizing the user scheduling policies in the underlying space-HAPS-ground networks, so as to determine the user-association strategy in an online fashion and highlight how such proposed method outperforms the traditional optimization approach. In light of the numerical prospects of the artificial-intelligence (AI)-based proposed solution, the paper also sheds light on the challenges and open issues that promise to spur the integration of machine learning in space-air-ground networks, namely, online HAPS power adaptation, learning-based channel sensing, data-driven multi-HAPSs resource management, and intelligent flying taxis-empowered systems. In light of the related works section above, the contributions of the current paper can be summarized as follows:}

\begin{itemize}
\item {The paper presents a concise overview of the existing machine learning-based methods that aim to solve the general resource allocation problems in existing wireless networks.}
\item {The paper addresses the problem of user scheduling in space-HAPS-ground networks subject to user-connectivity, backhaul, and power constraints, by maximizing the network throughput at the ground-level users.}
\item {The paper determines the user scheduling policy (i.e., user-to-HAPS or user-to-ground base-station) by using ensembling deep neural network for optimizing the user scheduling policies in an online fashion. }
\item {The paper simulations results show how the proposed ensembling deep neural network approach always outperforms the other classical offline optimized schemes, especially those recently developed in \cite{ShashaTWC2022}. The proposed solution particularly outperforms all other offline optimization solutions when the ensemble size increases, which illustrates the promising role of the proposed solution in optimizing future integrated space-air-ground networks.}
\item {The paper proposes a handful of timely open issues, which promise to spur the integration of machine learning in space-air-ground networks, namely, online HAPS power adaptation, learning-based channel sensing, data-driven multi-HAPSs resource management, and intelligent flying taxis-empowered systems.}
\end{itemize}
\subsubsection*{Organization}
The rest of the paper is organized as follows. Section \ref{ML_EDNN_SHAPSG} presents the applications of machine learning in the context of general resource allocation problems, and illustrates the prospects of using ensembling deep neural networks in optimizing the user scheduling in space-HAPS-ground systems. Section \ref{Future_Work} highlights the major challenges and open issues of using machine learning in future space-air-ground networks. The paper conclusions are finally presented in Section \ref{concl}.

\section{Machine Learning for Resource Allocation: The Case of Space-HAPS-Ground Networks}
\label{ML_EDNN_SHAPSG}
{Machine learning has recently emerged as a powerful solver for optimizing wireless communications \cite{zhu2020application, dahrouj2021overview}.
Thanks to its inherent characteristics in terms of data driven, automating analytical modeling, and online processing, machine learning proves to be useful at solving complex communication optimization problems with random, dynamic channel and unpredictable users demands in time and space \cite{dahrouj2021overview}. For the sake of completeness, we next briefly present how supervised, unsupervised, and reinforcement learning methods tackle the resource allocation problems in wireless systems.}

\subsection{Supervised Learning}
{When supervised learning is applied to wireless networks resource allocation problems, labeled data hinges upon deriving specific optimization algorithms, and then learning (approximating) the underlying function of the optimization algorithm. Such approach uses the fact that specific machine learning techniques, e.g., deep learning, can act as a universal function approximator; see
\cite{dahrouj2021overview} and \cite{sun2018learning} and references therein.}

\subsection{Unsupervised Learning}
{Unlike supervised learning which produces an approximate solution that is at best on the same par with the original algorithm, unsupervised learning directly uses the objective function as the loss function of the underlying wireless communication optimization problems, the objective functions and constraints of which are typically complex and non-convex. By further using specific penalty terms to account for the constraints, ensembling learning can also improve the system performance \cite{liang2019towards}, which is adopted in this paper in the context of user scheduling in VHetNets.}

\subsection{Reinforcement Learning}
{Reinforcement learning is an alternative approach that maximizes cumulative rewards. More specifically, reinforcement learning consists of agent, environment, state, action and reward. After the agent takes an action, the environment transitions to a new state and evaluates the reward. Subsequently, according to the new state and reward, the agent takes a new action and policy, in an effort to eventually maximize the long-term cumulative reward. Such approach is particularly useful in resource management problems that can be formulated as Markov decision processes, e.g., see \cite{cui2021scalable} and references therein.}\\

\subsection{Adopted Approach}
{As mentioned earlier, the current paper adopts an unnsupervised ensembling deep neural network approach to solve the user scheduling in VHetNets. The rationale behind adopting such an unsupervised learning algorithm as opposed to other machine learning types is that supervised learning is a rather suboptimal solution that can yield a solution which is at best as good as the known heuristics. The setup of the problem under study also does not fit the general framework of reinforcement learning, since the optimization problem under study does not assume a cumulative reward-based objective. Given how complex the discrete scheduling problem under study is-- which makes finding the global optimal solution of exponential complexity-- and given the stochasticity of the VHetNets underlying channel, the paper uses ensembling deep learning approach to solve the problem. The compelling feature of the proposed approach is that it combines the capabilities of the deep learning algorithm with the enhanced search space obtained through the ensembling approach.}

\subsection{User Association in Satellite-HAPS-Ground Systems Using Ensembling Deep Neural Networks (EDNN)}
This section considers a specific satellite-HAPS-ground network, and illustrates its potential at augmenting the ground-level communication, especially for connecting the unconnected and ultraconnecting the connected. In an effort to boost the digital inclusion index, the paper considers solving the user scheduling problem (i.e., user to geo-satellite via the HAPS, or user to BS) by imposing a specific user connectivity constraint. In fact, compared to conventional standalone terrestrial networks, scheduling problems in space-HAPS-ground networks are more complex \cite{ShashaTWC2022}. This mainly stems from the several modes of cross-layer interference of the considered system (i.e., intra-HAPS interference, intra-BS interference, inter-BS interference, and HAPS-BS interference), in addition to the stochasticity of their underlying channels. Given the random nature of the underlying space-air-ground channels, the paper, therefore, adopts an artificial-intelligence based technique to maximize the network sum-rate subject to specific connectivity and power constraints.
Ensembling deep neural networks are particularly adopted to solve such the user scheduling problem in an online fashion.
The simulation results then illustrate how the proposed solution acts a powerful enabler to improve the performance of the considered integrated satellite-HAPS-ground network.
\vspace{-0.5cm}
\subsection{The Considered System Model}
Consider an integrated satellite-HAPS-ground network, which consists of one geo-satellite, one HAPS and several ground base-stations (BSs), where the geo-satellite is connected to the HAPS via a free-space optical (FSO) link. The HAPS then transmits data to the ground users via radio-frequency (RF) links. On the ground level, the BSs which are equipped with multiple antennas serve users through spatial multiplexing, i.e., beamforming.
In addition, the gateway and satellite transmit data at different time intervals, and so we do not consider the interference between uplink and downlink.
An example of the considered network is illustrated in Fig. \ref{SM}. In fact, such networks have recently attracted attention in the context of 6G research directions.
From a link level analysis perspective, reference \cite{saeed2021P2P} investigates the point to point (P2P) links between the different entities of the three-layer network, i.e., space-air-ground networks. The characteristics, feasibility, advantages, and challenges of HAPS as air-based networks are also analyzed in \cite{alam2021high}, which highlights how HAPS quasi-stationarity makes it suitable for augmenting ground-level communications infrastructures. More recently, the results in \cite{ShashaTWC2022} prove that the integrated satellite-HAPS-ground network can majorally help mitigating the digital divide problem through joint user scheduling and beamforming. The discrete user scheduling solution provided in \cite{ShashaTWC2022}, however, relies on traditional optimization heuristics. To this end, the current paper goes one step forward in the direction of  developing machine-learning based-resource allocation solutions, and illustrates how ensembling deep neural networks-based solution outperforms the conventional systematic approach.

\vspace{-0.5cm}
\subsection{System Objective and Constraints}
\begin{figure}[!t]
\centering
\includegraphics[width=3in]{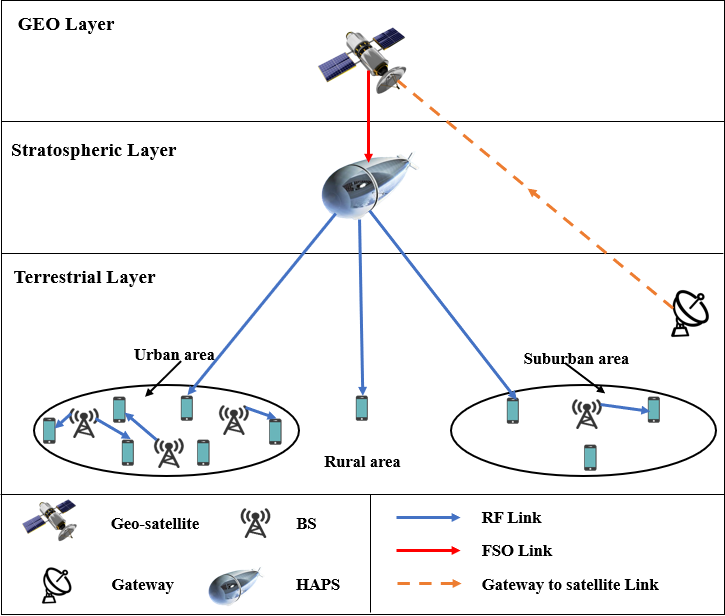}
\caption{System model.}
\label{SM}
\end{figure}
In this paper, we consider an integrated satellite-HAPS-ground network, which consists of one satellite, one HAPS, $N_{B}$ ground BSs and $N_{U}$ users. We denote the set of BSs by $\mathcal{B}=\{0,1,2,...,N_B\}$, where the $0^{th}$ BS is the HAPS, and others indices denote the ground BSs. Furthermore, we assume that $i^{th}$ BS is equipped with $N_{A}^{i}$ antennas, and we denote the set of users by $\mathcal{U}=\{1,2,...,N_U\}$. The satellite and HAPS are connected via FSO links, and each user is served by either one of the BSs or by the HAPS via RF links. The gateway and satellite transmit data at different time intervals, so we do not consider the interference between uplink and downlink. An example of the considered network is illustrated in Fig. \ref{SM}. The paper connectivity constraints impose that each user can only be served by at most one BS or by the HAPS, and that the number of users served by HAPS or BSs can not exceed their respective payloads. The transmissions from the HAPS to users and BSs to users use the same RF bands, but adopt beamforming to serve multiple users simultaneously.

Let $\alpha_{ij}$ be the binary association variable of user $j$ to BS $i$. Also, let $\mathbf{h}_{ij}\in\mathbb{C}^{N_{A}^{i}}$ be the channel between user $j$ and BS $i$, $\mathbf{w}_{ij}$ be the beamforming vector of user $j$ when served by BS $i$, $\sigma^{2}$ be the noise power, and $\gamma_{ij}$ be the availability binary variable of user $j$'s signal at BS $i$, $\forall i\in \mathcal{B}$. The rate of user $j$, once served by BS $i$, can, therefore, be written as:
 \begin{footnotesize}
\small
\begin{equation}
\label{R_ij}
R_{ij}^{Ground\_BS}\!= \!\beta\log_{2}\left(1 \!+ \!\frac{|\mathbf{h}_{ij}^{H}\mathbf{w}_{ij}|^{2}}{\sum_{u=1,u\neq j}^{N_{U}}\sum_{b=0}^{N_{B}}\gamma_{bu}\alpha_{bu}|\mathbf{h}_{bj}^{H}\mathbf{w}_{bu}|^{2} + \sigma^{2}}\right),
\end{equation}
\end{footnotesize}

\noindent where $\beta$ is the bandwidth. Similarly, the achievable rate of user $j$ served by the HAPS via the RF link can be written as:
\begin{footnotesize}
\small
\begin{equation}
\label{R_0j}
R_{0j}^{HAPS\_RF}=\beta\log_{2}\left(1+\frac{|\mathbf{h}_{0j}^{H}\mathbf{w}_{0j}|^{2}}{\sum_{u=1,u\neq j}^{N_{U}}\sum_{b=0}^{N_{B}}\gamma_{bu}\alpha_{bu}|\mathbf{h}_{bj}^{H}\mathbf{w}_{bu}|^{2} + \sigma^{2}}\right).
\end{equation}
\end{footnotesize}

This paper focuses on maximizing the sum-rate subject to user-connectivity, backhaul, and power constraints, so as to determine $\alpha_{ij}$. The problem can then be written as follows:

\begin{subequations}
\label{EHM}
\begin{eqnarray}
\label{EHMa}
&\displaystyle\max_{\alpha_{ij}}&  \sum_{j=1}^{N_{U}}\left(\sum_{i=1}^{N_{B}}\gamma_{ij}\alpha_{ij}R_{ij}^{Ground\_ BS}\!+\!\gamma_{0j}\alpha_{0j}R_{0j}^{HAPS}\right),\\
\label{EHMb}
&s.t.& ( \ref{R_ij}), (\ref{R_0j}),\\
\label{EHMd}
&& R_{0j}^{HAPS}=\min\{R_{0j}^{HAPS\_RF}, R_{FSO}\},\\
\label{EHMe}
&&  \sum_{j=1}^{N_{U}}\gamma_{ij}\alpha_{ij} \leq M_{i},\ \forall i \in \mathcal{B},\\
\label{EHMf}
&& \sum_{i=0}^{N_{B}}\gamma_{ij}\alpha_{ij} \leq 1,\  \forall j \in \mathcal{U},\\
\label{EHMg}
&& \sum_{i=0}^{N_{B}}\sum_{j=1}^{N_{U}}\gamma_{ij}\alpha_{ij} \geq K,\\
\label{EHMh}
&& \alpha_{ij}\in \{0, 1\},\  \forall i \in \mathcal{B}, \forall j \in \mathcal{U},\\
\label{EHMi}
&&\sum_{j=1}^{N_{U}}\gamma_{ij}\alpha_{ij}\mathbf{w}_{ij}^{H}\mathbf{w}_{ij}\leq P_{i}^{max},\ \forall i \in \mathcal{B},
\end{eqnarray}
\end{subequations}
\noindent where the optimization is over the variables $\alpha_{ij}$, constraint (\ref{EHMd}) accounts for the FSO link capacity, $ M_{i}$ is the payload of BS $i$, $K$ is the minimum number of users to be served, and $P_{i}^{max}$ denotes the maximum power of BS $i$, $\forall i\in \mathcal{B}$. It is worthwhile to note that the classical approach to solve problem (\ref{EHM}) often relies on solving a snapshot of the problem for fixed channel values, and repeating the process whenever the channel changes. Such process, however, requires the algorithmic processing time to be less than the time of channel changes. Obviously, this is not always possible with model-based traditional algorithms. Therefore, our paper next proposes using ensembling neural networks to achieve real-time processing and gain better performance.

\subsection{Ensembling Deep Neural Networks Approach}
Since supervised learning is akin to a redesign of existing traditional methods (through approximating complex algorithms), this paper rather adopts unsupervised learning to address the user scheduling dilemma. The paper particularly uses ensembling deep neural networks (EDNN) to solve the problem described above, due to EDNN promising numerical prospects; see \cite{liang2019towards} and references therein. Using an online optimization approach, the maximization problem (\ref{EHM}) can be recast as a minimization of the following loss function:
 \begin{subequations}
\label{EHM1}
\begin{eqnarray}
\label{EHM1a}
&\displaystyle L&
 =\mathbb{E}[-R(\mathbf{H},\gamma,\boldsymbol{\theta})
         \! +\!(\sum_{j=1}^{N_{U}}\mu_{j}C_{1,j})
           \! +\!(\sum_{i=0}^{N_{B}}\lambda_{i}C_{2,i}) \nonumber\\
             & &
             \! +\!(\sum_{i=0}^{N_{B}}\delta_{i}C_{3,i})
             \! +\! \rho C_{4}],\\
\label{EHM1b}
&\text{where}&   C_{1,j}=ReLU(\sum_{i=0}^{N_{B}}\gamma_{ij}\alpha_{ij}-1),\\
\label{EHM1c}
&& C_{2,i}=ReLU(\sum_{j=1}^{N_{U}}\gamma_{ij}\alpha_{ij}-M_{i}),\\
\label{EHM1d}
&&C_{3,i}=ReLU(\sum_{j=1}^{N_{U}}\gamma_{ij}\alpha_{ij}\mathbf{w}_{ij}^{H}\mathbf{w}_{ij}-P_{i}^{max}),\\
\label{EHM1e}
&&C_{4}=ReLU(K-\sum_{i=0}^{N_{B}}\sum_{j=1}^{N_{U}}\gamma_{ij}\alpha_{ij}),
\end{eqnarray}
\end{subequations}
where $ReLU(.)$, i.e., the rectified linear activation function, accounts for the constraints of problem (\ref{EHM}).

\begin{figure}[!t]
\centering
\includegraphics[width=3in]{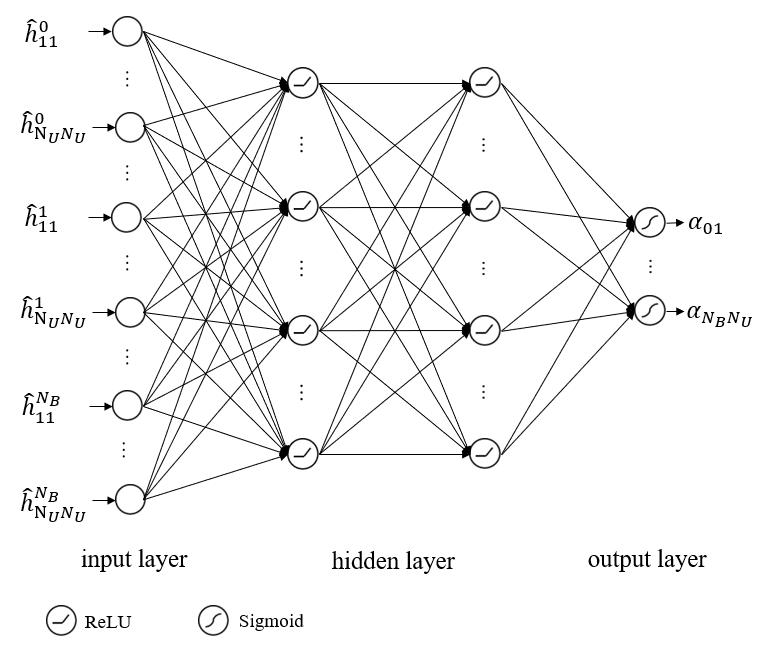}
\caption{The structure of deep neural network (DNN).}
\label{DNN}
\end{figure}

Minimizing the loss function (\ref{EHM1a}) takes the system parameters $\mathbf{h}_{ij}$ and $\mathbf{w}_{ij}$ as inputs, and outputs the association variables $\alpha_{ij}$. Our paper proposes a fully connected deep neural network to solve the user scheduling problem (\ref{EHM}). More specifically, the neural network consists of $2$ hidden layers which have $ReLU$ activation functions and adopts batch normalization. Because the value of $\alpha_{ij}$ is between $0$ and $1$, the output layer has a sigmoid activation function. The structure of the deep neural network is shown in Fig. \ref{DNN}.
It is worth mentioning that in order to satisfy the constraints of the problem (\ref{EHM}), we use the Lagrangian dual ascent to update the regularization parameters. The overall algorithm are shown in Algorithm \ref{G1} description below.
\begin{algorithm}[h!]
 \caption{Overall Algorithm}
 \label{G1}
 \begin{enumerate}
  \item Initialize $\lambda_{i}, \delta_{i}, \mu_{j}, \rho$,
  \item Apply neural netwok to get $\alpha_{ij}^{k+1}=\arg\min_{\alpha_{ij}}L(\alpha_{ij}, \lambda_{i}^{k}, \delta_{i}^{k}, \mu_{j}^{k}, \rho^{k})$.
   \item $\mu_{j}^{k+1}=\mu_{j}^{k}+\beta\frac{\partial L}{\mu_{j}}=\mu_{j}^{k}+\beta\mathbb{E}(C_{1,j})$.
  \item $\lambda_{i}^{k+1}=\lambda_{i}^{k}+\beta\frac{\partial L}{\lambda_{i}}=\lambda_{i}^{k}+\beta\mathbb{E}(C_{2,i})$.
  \item $\delta_{i}^{k+1}=\delta_{i}^{k}+\beta\frac{\partial L}{\delta_{i}}=\delta_{i}^{k}+\beta\mathbb{E}(C_{3,i})$.
  \item $\rho^{k+1}=\rho^{k}+\beta\frac{\partial L}{\rho}=\rho^{k}+\beta\mathbb{E}(C_{4})$.
  \item Stop at convergence.
 \end{enumerate}
 \end{algorithm}

\subsubsection{Data Generation}
The channel value between user $j$ and BS $i$
$\mathbf{h}_{ij}\in\mathbb{C}^{N_{A}^{i}}$ following certain distribution can be found in \cite{ShashaTWC2022}, which accounts for path loss, shadowing, and multi-path. Furthmore, we fix the beamforming vectors of each ground BSs and HAPS, noise power $\sigma^{2}$, availability binary variable $\gamma_{ij}$ ($\forall i\in \mathcal{B}, \forall j\in \mathcal{U}$), and the location of users.
To reduce the dimension, the input to the network is the effective channels between BSs and users, which can be represented as $\hat{\mathbf{h}}_{lj}^{i}=|\mathbf{h}_{il}\mathbf{w}_{ij}|^{2}$ for each BS or HAPS, which can be interpreted as the interference comes from user $j$ served by BS $i$ ($l\neq j$) for user $l$ and receive signal of user $l$ served by BS $i$ ($l=j$). Furthermore, $\mathbf{H}$ is the feature vector which is flattened to contain corresponding $\hat{\mathbf{h}}_{lj}^{i}$.

\subsubsection{Training Stage}
The entire training data set is used to optimize the weights of the neural network associated with the loss function (\ref{EHM1a}). In this paper, we use Xavier initialization \cite{glorot2010understanding} to initialize the network weights and Adam optimizer \cite{kingma2014adam} to train the network. The mini-batch stochastic gradient descent is then used to calculate the gradient. During training, a validation data set is used to evaluate the performance of the neural network.

In order to further improve upon the solution reached by a single neural network solution, the paper adopts training an ensemble of networks, and choosing the best resulting solution, similar to the approach used in \cite{liang2019towards}. Note that for each of the ensembling of deep neural networks, they are trained with different training data and initialization methods.

\subsubsection{Testing Stage}
We generate a test set with the same channel distribution as the training dataset. For every channel realization, we pass it into the trained ensembling networks, select the network which can produce the best sum-rate and collect the optimized user scheduling scheme.

To assess the numerical prospects of the above solution, the paper also simulates some classical discrete optimization methods, namely, the integer linear programming and generalized assignment problem (ILP-GAP), the channel dependent (CD), and distance dependent (DD) methods as baselines, the details of which can be found in \cite{ShashaTWC2022}.

\subsection{Simulations}
\begin{figure}[!t]
\centering
\includegraphics[width=3in]{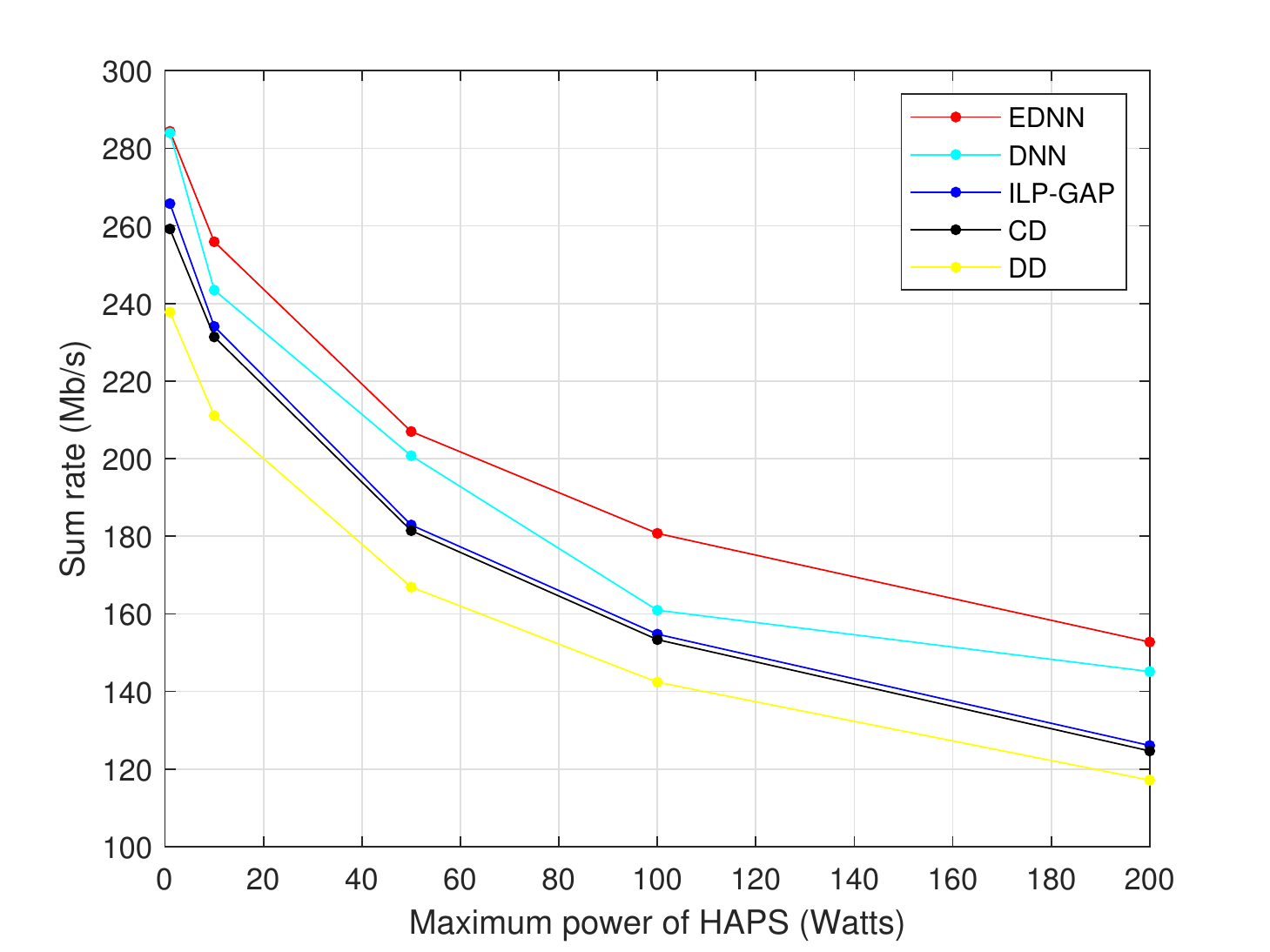}
\caption{Sum-rate versus maximum power of HAPS.}
\label{HAPS_power}
\end{figure}
This section illustrates the performance of the proposed EDNN-based solution versus the maximum power of HAPS, the $K$-value (i.e., the minimum number of users to be served.), the ensembling size and the running time, so as to highlight the numerical prospect of the above solution. We consider an integrated satellite-HAPS-ground network, wherein $20$ users are distributed in three different subareas. Subarea $1$ contains $6$ BSs and $12$ users. Subarea $2$ contains $3$ BSs and $6$ users. The remaining area is Subarea $3$ and contains $2$ users, with $1$ BS.
The HAPS is equipped with $10$ antennas.
\subsubsection{Parameter Setting}
In addition to the parameters of the communication network mentioned above, other parameters in the communication network can be found in reference \cite{ShashaTWC2022}.
About the hyper-parameters of the deep neural network, unless specified otherwise, the mini-batch size and learning rate are taken as $50$ and $0.001$. For every iteration, the number of epochs is $150$. Moreover, initial penalty parameters $\lambda_{i}, \delta_{i}, \mu_{j}, \rho$ is $0.1$ and the step size $\beta$ of updating the penalty parameters is set to $1$. For every single deep neural network (DNN), the number of neurons from input layer to the output layer are given as $\{600, 1200, 600, 220\}$. For ensembling deep neural networks (EDNNs), the ensembling size is set to be $8$. the training data size and testing data size are $5000$ and $500$.
\subsubsection{Numerical Result}
Firstly, Fig. \ref{HAPS_power} shows the sum-rate versus the maximum power of HAPS when $K$ is $15$. It is clear that, as the maximum power of HAPS increases, all highlighted solutions decrease, due to the increased interference levels. Fig. \ref{HAPS_power} particularly shows how the proposed (online) deep learning approach always outperforms the other classical offline optimized schemes developed in \cite{ShashaTWC2022}. The EDNN-based solution particularly outperforms all other simulated solutions for all values of the HAPS power, which highlights the numerical prospect of advanced machine leaning techniques in solving sophisticated resource allocation problems.

\begin{figure}[!t]
\centering
\includegraphics[width=3in]{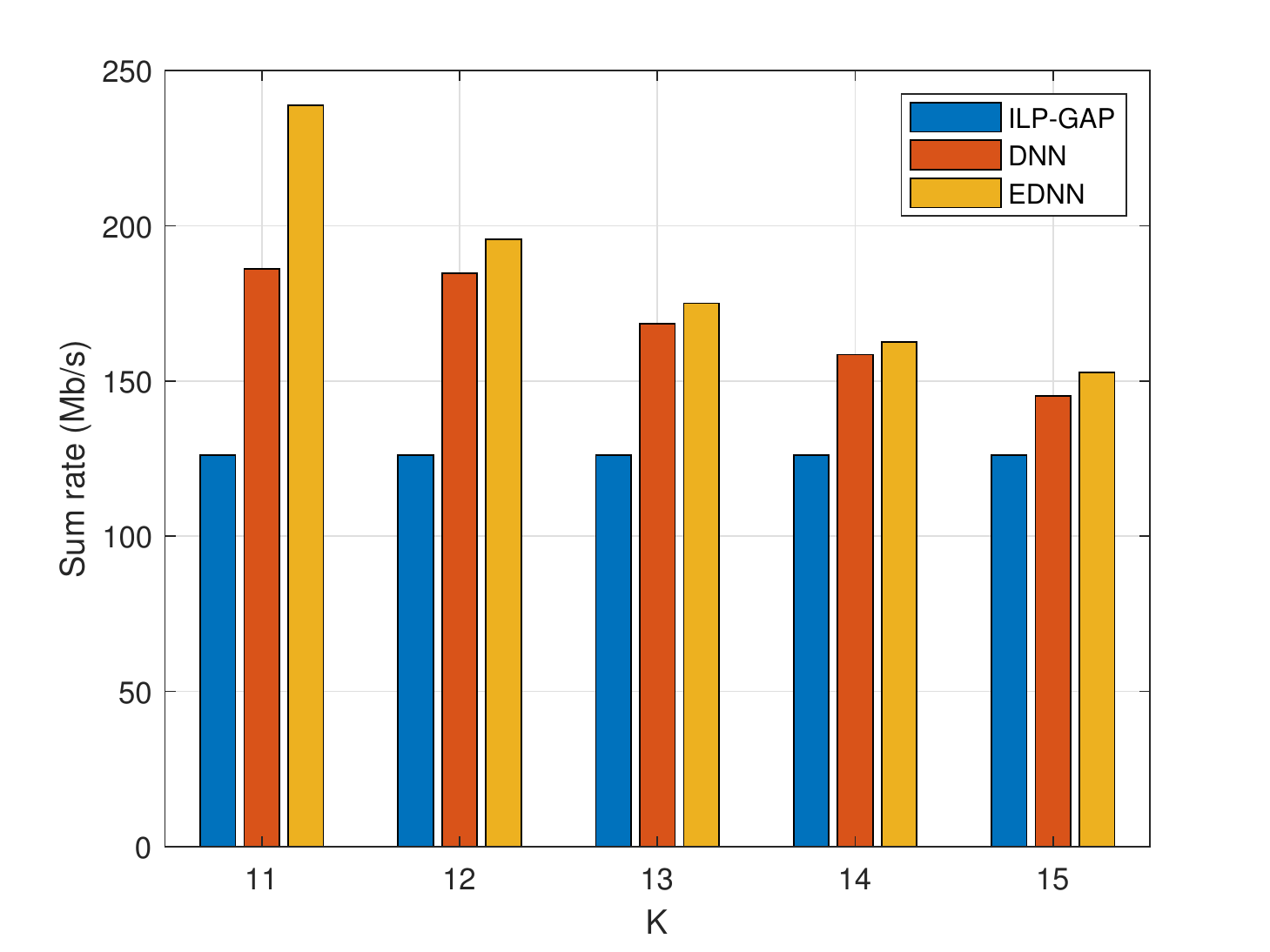}
\caption{Sum-rate versus values of $K$.}
\label{K value200}
\end{figure}
To illustrate the impact of the proposed solution in connecting the unconnected and ultraconnecting the connected, Fig. \ref{K value200} shows a comparison of ILP-GAP versus the deep neural network (DNN)-based algorithms for different values of $K$ when the maximum power of HAPS is $200$Watts. The figure shows that, as the value of $K$ increases, the sum-rate of the proposed DNN approach decreases, due to more users served by the system, which introduces more interference. The figure, however, shows that DNN approach always has a superior performance as compared to the offline optimization approach.

\begin{figure}[!t]
\centering
\includegraphics[width=3in]{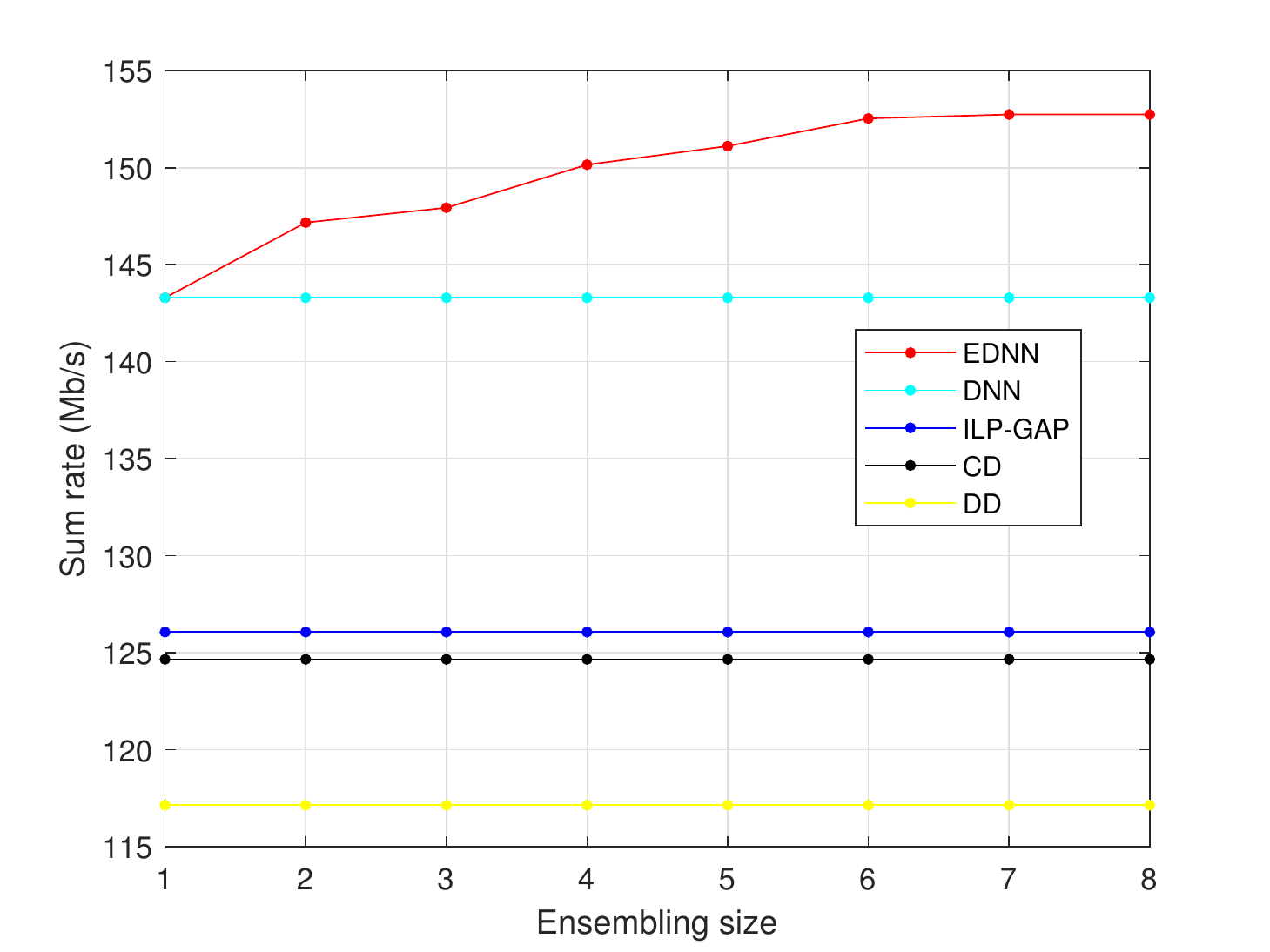}
\caption{Sum-rate versus ensembling size.}
\label{I_200}
\end{figure}
{Fig. \ref{I_200} represents the added value for using ensembling learning when the maximum power of HAPS is set to $200$Watts and $K$ is set to $15$. Fig. \ref{I_200} shows that, with the ensembling size increasing, the sum-rate also increases. This is because the considered user scheduling problem is a discrete non-convex problem, which typically has several local optima. Therefore, DNN with different training data and initialization methods improves the local optima search, and can reach up to 26\% in terms of sum-rate as compared to the classical offline optimized schemes developed in \cite{ShashaTWC2022}.
Fig. \ref{I_200} further reemphasizes how the proposed machine learning-based techniques always outperform all the offline optimized methods for all values of the ensembling size.}

\begin{table*}[!t]
\centering
\caption{Sum-rate and computational time performance}
\label{table-time}
\begin{tabular}{|p{.2\textwidth} | p{.2\textwidth} | p{.2\textwidth} |  }
\hline
  \textbf{Schemes} &  \textbf{Sum-Rate Performance (Mb/s)}&   \textbf{Running Time (ms)} \\
 \hline
ILP-GAP &123.451&1082.654 \\
  \hline
DNN&  141.778&23.970\\
  \hline
EDNN (with ensembling size $8$)&  150.210&26.060\\
 \hline
\end{tabular}
 \end{table*}
Finally, to illustrate the computational complexity and sum-rate performance of the different association methods, Table \ref{table-time} depicts the sum-rate performance, time complexity, and training time, when $K$ is $15$ and the maximum power of HAPS is $200$Watts. In this part of the simulations, DNN and EDNN are implemented in python 3.8.0 with Torch  1.8.0 on one computer (Intel Core-i5 processor). Table \ref{table-time} illustrates that the  machine learning-based approaches (i.e., DNN and EDNN) can achieve higher sum-rate with lower computational complexity. In particular, note that our proposed DNN and EDNN methods outperform the ILP-GAP solution by a significant margin of $1058.684$ms and $1056.594$ms, respectively, which provides an additional valuable testimony of the proposed solution in the context of optimizing future integrated space-air-ground networks.

\section{Challenges and Open Issues}
\label{Future_Work}
{Despite the numerical advantages of artificial intelligence-based methods in integrated space-air-ground networks, their practical implementation remains a strong function of system challenges, such as spatio-temporal channel variations, scarcity of dataset, heterogeneity, and data collection hurdles}. On the one hand, the current hardware, including batteries and on-system components, cannot meet the stringent requirements of integrated space-air-ground networks. On the other hand, despite its great potential, machine learning relies heavily on the data availability and is computationally demanding at the training stage. To this end, this section sheds light on some of the noticeable challenges and presents some promising open issues on this topic.
\subsection{Online HAPS Power Adaptation}
Flight control and communication payload are among the major aspects of HAPS energy consumption \cite{kurt2021vision}.
Specifically, HAPS flight control often imposes quasi-stationary constraints, which necessitates propulsion power.
HAPS can also be often treated as a super-macro base-station, the energy consumption of which includes transmiting and processing information signals. Since the height of HAPS is generally in the order of $18$km-$21$km \cite{alam2021high}, the considerable path loss requires to be compensated with high transmit power.
Unlike ground BSs which can be connected to the electrical grid, however, HAPS operation relies on the available on-board energy sources. So far, there are three types of energy sources that serve to empower HAPS (i.e., battery energy source, energy beams, and laser beams) \cite{kurt2021vision}. While conventional battery energy cannot provide a long HAPS endurance, the other two energy sources cause high power irradiation risks. To this end, one considerable way to increase the energy of HAPS is to use solar or hydrogen fueled power, especially given HAPS positioning at the stratosphere. Given the ever-changing spatio-temporal conditions of the surrounding environment, the problem of dynamically choosing the proper source to empower HAPS (i.e., battery energy source, energy beams, laser beams, solar or hydrogen) becomes innately related to adopting the most appropriate learning-based approaches, which promises to be a timely topic for future investigation.
\subsection{Learning-Based Channel Sensing}
Channel knowledge is a key factor in optimizing future integrated space-air-ground communications. Given the ever-changing users' locations at the ground level, and the channel random parameters (i.e., shadowing, multi-path, etc.), the perfect channel state information knowledge would require a perplexed process that is impossible to realize in real-time. Such process is particularly intricate in the complex integrated multi-layered terrestrial and non-terrestrial networks. Invoking learning-based channel
sensing methods to excavate the channel state information through geographic location information, weather conditions, etc., is, therefore, a timely future research direction, which promises to intertwine statistical signal processing techniques with both distributed and centralized machine learning tools.
\subsection{Data-Driven Resource Management in Multi-HAPSs Scenarios}
Multi-HAPSs systems are expected to provide a distributed air-platform enabler to achieve global seamless connectivity \cite{kurt2021vision}. Under such systems, however, several cross-layer interference modes are bound to coexist, namely, intra-HAPS interference, inter-HAPS interference, intra-BS interference, inter-BS interference, and HAPS-BS interference. The resource management problems in multi-HAPSs scenarios become, therefore, more complex than its one-HAPS scenario counterpart. In essence, the approach proposed in section III assumes that decisions are made within a centralized processor, i.e., through centralized learning which requires that all collected data need to be uploaded to the centralized processor; thereby raising a multitude of security and privacy concerns. To best handle resource allocation problems in multi-HAPS scenarios, federated learning, where each data owner does not need to share their own data set, promises to offer an efficient paradigm to enable a democratized distributed learning platform \cite{dahrouj2021overview}. In fact, the marriage of ensembling deep learning with federated learning can play a major role in tackling the intricacy of security and privacy concerns through multi-HAPS data-driven resource management protocols, which promises to be an active area for future research.
\subsection{Intelligent Flying Taxis-Empowered Systems}
Flying vehicles (e.g., flying taxis, delivery systems, etc.) are regarded as the potential futuristic solution to handle the traffic congestion and the rapid expansion of the delivery industry \cite{pan2021flying}. Augmenting future intelligent transportation systems (ITS) with machine learning capabilities are expected, therefore, to spark major research fields for boosting flying taxis-empowered systems, especially under the framework of future smart cities. More specifically, flying vehicles require dynamic control signals to avoid collisions, overcrowding, etc. Data transmission across vehicles, further, would guarantee voice and multimedia high quality-of-service. Given the height of future flying vehicles above the ground levels, the high-altitude platforms can help enhancing both their control signals and their quality-of-service targets. Considering the contextual mobility across the flying cars, in addition to the handover and channel variations, investigating data-driven resource allocation schemes in integrated HAPS-flying taxis systems becomes essential in the context of next generation ITS design. Such emerging AI-empowered interdisciplinary research direction that falls at the intersection of communication, transportation, and computation, promises to be a timely topic for future investigation.
\vspace{-0.5cm}
\section{Conclusion}
\label{concl}
{ The integration of artificial intelligence into future communication networks is expected to spur a myriad of possible advantages in the context of space-air-ground communications. To this end, this paper goes one step forward toward  developing machine-learning based-resource allocation solutions in satellite-HAPS-ground networks. The paper first overviews the most relevant state-of-the art in the context of machine learning applications to the resource allocation problems in space-air-ground networks. The paper then proposes, and shows the benefit of, one specific application that uses ensembling deep neural network for optimizing the user scheduling in integrated space-HAPS-ground networks. The proposed solution is particularly shown to outperform all other offline optimization solutions when the ensemble size increases (with up to 26\% in sum-rate improvement), which illustrates the promising role of the proposed solution in optimizing future integrated space-air-ground networks. The paper finally presents a handful of challenges and open issues that promise to spur the integration of machine learning in space-air-ground networks, e.g., online HAPS power adaptation, learning-based channel sensing, data-driven multi-HAPSs resource management, and intelligent flying taxis-empowered systems-- topics that are soon expected to trigger the discussion on beyond 6G systems.}

\bibliography{my_bibliography}
\bibliographystyle{IEEEtran}
\end{document}